\documentclass[letterpaper]{article} 
\usepackage{arxiv}  
\usepackage{times}  
\usepackage{helvet}  
\usepackage{courier}  
\usepackage[hyphens]{url}  
\usepackage{graphicx} 
\usepackage{natbib}  
\usepackage{caption} 
\usepackage{algorithmic}

\usepackage{dsfont}
\usepackage{amsmath}
\usepackage{amssymb}
\usepackage{booktabs}
\usepackage{multirow}
\usepackage[ruled,vlined]{algorithm2e}
\usepackage{enumitem}

\newtheorem{myhypo}{\textsc{Hypothesis}}
\newcommand{\Hquad}{\hspace{0.5em}}
\usepackage{colortbl}
\definecolor{Gray}{gray}{0.9}
\makeatletter
\DeclareRobustCommand\onedot{\futurelet\@let@token\@onedot}
\def\@onedot{\ifx\@let@token.\else.\null\fi\xspace}

\def\eg{\emph{e.g}\onedot} 
\def\ie{\emph{i.e}\onedot} 
 
 \def\vs{\emph{vs}\onedot}
 
\def\etal{\emph{et al}\onedot}
\makeatother
\title{PointCVaR: Risk-optimized Outlier Removal for Robust 3D Point Cloud Classification}
\author {
    Xinke Li\textsuperscript{\rm 1, *},
    Junchi Lu\textsuperscript{\rm 2, *},
    Henghui Ding\textsuperscript{\rm 2},
    Changsheng Sun\textsuperscript{\rm 1},
    Joey Tianyi Zhou\textsuperscript{\rm 3, \rm 4},
    Chee Yeow Meng\textsuperscript{\rm 1}
}
\affiliations {
    \textsuperscript{\rm 1}National University of Singapore \Hquad
    \textsuperscript{\rm 2}Nanyang Technological University\\
    \textsuperscript{\rm 3}
Institute of High Performance Computing (IHPC), A*STAR \Hquad
    \textsuperscript{\rm 4}
Centre for Frontier AI Research (CFAR), A*STAR\\
   \{xinke.li, cssun\}@u.nus.edu\Hquad LUJU0008@e.ntu.edu.sg\\
henghui.ding@gmail.com\Hquad zhouty@cfar.a-star.edu.sg\Hquad ymchee@nus.edu.sg
}

\usepackage{bibentry}

\begin{document}

\maketitle

\begin{abstract}
With the growth of 3D sensing technology, the deep learning system for 3D point clouds has become increasingly important, especially in applications such as autonomous vehicles where safety is a primary concern. However, there are growing concerns about the reliability of these systems when they encounter noisy point clouds, either occurring naturally or introduced with malicious intent. This paper highlights the challenges of point cloud classification posed by various forms of noise, from simple background noise to malicious adversarial/backdoor attacks that can intentionally skew model predictions. While there's an urgent need for optimized point cloud denoising, current point outlier removal approaches, an essential step for denoising, rely heavily on handcrafted strategies and are not adapted for higher-level tasks, such as classification. To address this issue, we introduce an innovative point outlier removal method that harnesses the power of downstream classification models. Using gradient-based attribution analysis, we define a novel concept: \textit{point risk}. Drawing inspiration from tail risk minimization in finance, we recast the outlier removal process as an optimization problem, named \textit{PointCVaR}. Extensive experiments show that our proposed technique not only robustly filters diverse point cloud outliers but also consistently and significantly enhances existing robust methods for point cloud classification. A notable feature of our approach is its effectiveness in defending against the latest threat of backdoor attacks in point clouds.  
\end{abstract}

The accessibility of 3D sensing technology has led to the popularity of deep learning on 3D point clouds in various industrial applications, including safety-critical ones such as autonomous driving~\cite{chen2017multi}. However, concerns over the safety of 3D deep learning have been substantiated by recent studies that have demonstrated the negative impact of natural or artificial noisy points on point cloud deep model performance~\cite{xiang2019generating, sun2022benchmarking}. Consequently, the reliability of 3D deep visual systems in industry is now overshadowed.

\begin{figure}[!ht]
    \centering
    \includegraphics[width=0.95\linewidth]{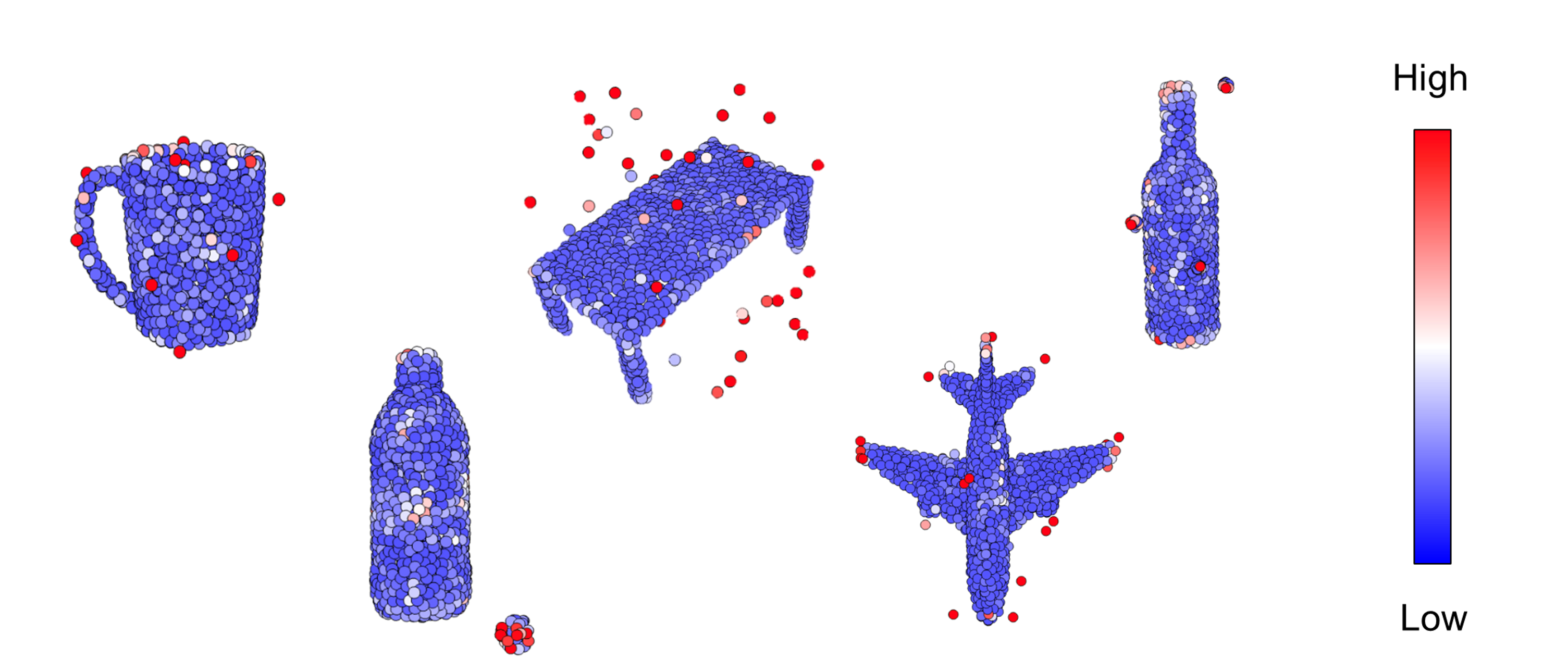}
    \caption{Visualizing point risks for point clouds with various types of outliers. It shows that the noise points often pose high risks compared to the clean points.}\label{fig:intro}
\end{figure}

Indeed, it is proposed that even some common types of point cloud noise could degrade the performance of deep models obviously. For instance, background points resulting from inaccurate sensors or incomplete processing can severely disrupt the predictions of point cloud classifiers, as revealed by~\cite{ren2022benchmarking}. Moreover, some artificial modifications to the points can be disguised as noise and intentionally manipulate the output of classifiers. On one hand, delicately designed adversarial noise, such as outliers or slight perturbations, has been shown to significantly decrease the classification accuracy in various studies~\cite{xiang2019generating,  wicker2019robustness}. On the other hand, a new form of attack, called a backdoor attack, has emerged as a severe threat. This attack involves implanting malicious functionality into a deep model by modifying the training data. During the inference of this model, if the input point cloud contains specific outliers named backdoor triggers, such as a small sphere, the model will output an incorrect target label~\cite{li2021pointba}. Thus, developing robust point cloud denoising pipelines becomes imperative to ensure the safety and reliability of 3D deep learning.

Despite the pressing need for point cloud denoising towards reliable deep learning, outlier removal, an essential processing step in a point cloud denoising pipeline, has not been specifically designed or optimized for deep learning yet. In current practice, the objective of outlier removal is eliminating points in the point cloud that do not conform to the geometrical characteristics of the local points. It is utilized to remove noise points directly or to pave the way for further filtering. Traditional methods, such as statistical outlier removal~\cite{rusu2010semantic}, have been demonstrated to be useful in preventing adversarial noise~\cite{zhou2019dup}. However, these methods heavily rely on handcrafted procedures and suffer from inferior performance for certain outliers, \eg, backdoor triggers~\cite{li2021pointba}. There have also been some learning-based attempts, such as PointCleanNet~\cite{rakotosaona2020pointcleannet}, which operates on a local patch of points. Although these methods are effective against specific types of noise, they incur non-trivial computational costs and greatly depend on the construction of noise datasets with ground truth. Besides their inherent limitations, the above outlier removal methods have not yet considered their impact on high-level tasks including classification, and thus, their use in the context of point-cloud classification may result in suboptimal performance.

To meet such needs, we propose an effective point outlier removal method in this work by leveraging the downstream deep models. Initially, we employ gradient-based attribution analysis to determine the point-wise sensitivity to the model output, referred to as \textit{point risk}. Further observation reveals that noise points mainly contribute to the tail risk in the point risk distribution, \ie, they are sparse but high in risk as showcased in Figure~\ref{fig:intro}. Motivated by tail risk minimization in financial applications~\cite{rockafellar2000optimization},\textbf{ we propose to formulate the outlier removal process into an optimization problem by introducing a convex risk measure.} Specifically, the process uses a tail risk measure as the objective and utilizes linear programming to obtain binary weights for input points, which are used to process the point cloud subsequently. The algorithm can be further approximated to linear computational complexity for higher efficiency. Extensive experiments demonstrate that the proposed method can effectively filter various types of point cloud outliers including random, adversarial, and backdoor trigger points, and its improvement in point cloud classification accuracy can exceed existing outlier removal methods. Moreover, our method can be combined with existing methods as a plugin to enhance the model's robustness to many adversarial noises.
In general, we claim to have made the following three contributions\footnote{Appendix and other resources are available on the public page: https://github.com/shinke-li/pointcvar }.
\begin{itemize}
    \item We analyze the noise within point clouds via the lens of deep model risk analysis, revealing that the noise points lead to elevated tail risk in the point risk distribution.
    
\item We derive an explainable outlier removal method for point clouds from the tail risk optimization, which only uses a trained classification model and thus avoids extra training or model modifications. It is also compatible with existing methods for robust point cloud classification.

\item Extensive experiments of point cloud classification show that the designed method can effectively remove various types of point outliers. Especially,  this method is, as far as we know, the first to successfully defend against point cloud backdoor attacks. Furthermore, it can consistently enhance existing robust techniques for 3D deep models when utilized as a plugin.
\end{itemize}

\section{Related Work}

\subsection{Noise Effect on Point Cloud Classification}
Deep learning models have shown impressive results in 3D point cloud classification~\cite{qi2017pointnet,xu2021learning}. Yet, they struggle with noisy data. Real-world point clouds often contain noise types that affect classifiers' performance~\cite{uy2019revisiting,ren2022benchmarking,sun2022benchmarking}. There are also adversarial noise attacks, including point perturbation~\cite{xiang2019generating,liu2019extending,tsai2020robust,hamdi2020advpc}, adversarial point addition~\cite{xiang2019generating,wicker2019robustness,yang2019adversarial}, and varied real-world noise simulations~\cite{zheng2019pointcloud,zhao2020isometry}. New backdoor attacks have been introduced that train models to recognize specific noise or triggers~\cite{li2021pointba, xiang2021backdoor}. Outlier removal, a preprocessing technique, can reduce noise impact on classification~\cite{zhou2019dup}. It aims to remove points disrupting point cloud uniformity using techniques like SOR~\cite{rusu2010semantic} and PointCleanNet~\cite{rakotosaona2020pointcleannet}. We aim to improve outlier removal efficacy by leveraging classification information.

\subsection{Robust Point Cloud Classification}
Three types of methods have been proposed to improve robustness of deep point cloud classifiers. \textbf{Model-based} methods focus on enhancing the model via designs or training strategies, such as the gather module by Dong~\etal and sorting-based pooling operations by Sun~\etal \cite{dong2020self, sun2020adversarial}. Robust training is also achieved through point cloud data augmentation methods such as PointMixup, PointCutMix, and adversarial training \cite{chen2020pointmixup, zhang2022pointcutmix, sun2021adversarially}. \textbf{Certified} methods, like the one by Liu~\etal, theoretically increase robustness to noise through certified classification, despite high computational costs \cite{liu2021pointguard}. \textbf{Data-based} methods aim to retrieve clean point cloud data from the noisy version using techniques like DUP-Net and IF-defense \cite{zhou2019dup, wu2020if}. Developing more effective outlier removal methods could potentially enhance the robustness of these techniques, especially data-based methods, as they directly involve outlier removal.

\subsection{Attribution Analysis on Deep Learning}\label{sec:section3}
With the rapid development of deep learning, dozens of deep model attribution analyses were proposed, such as Grad-CAM~\cite{springenberg2015striving}, Integrated Graients~\cite{sundararajan2017axiomatic} and Deep LIFT~\cite{shrikumar2017learning} on 2D image data. A few works explored this analysis in 3D point cloud through techniques of saliency points~\cite{zheng2019pointcloud,wicker2019robustness} and Shapley values~\cite{shen2021interpreting}.  In this work, we use attribution analysis as a tool to characterize the noisy samples and also empower our method with explainability.
\section{Risk Analysis }
We first introduce the concept of point risk. Based on a tail risk measure, we then indicate that noise points could result in a higher level of tail risk in the distribution of point risk.

\subsection{Point Risk Analysis} 
\textbf{Point Risk.} Given a point cloud $\mathcal{P}=\{\mathbf{x}_i\}_{i=1}^{N}$ for $\mathbf{x}_i\in\mathbb{R}^{3}$, a point cloud classifier $f(\cdot)$ can predict the label of $\mathcal{P}$ by outputting the label scores denoted by $f(\mathcal{P})$. Specifically, let $f(\cdot)$ be a composite function with $L$ intermediate layers, \ie, $f(\mathcal{P}) = f^{(L)}\left(\cdots f^{(2)}\left(f^{(1)}\left(\mathcal{P}\right)\right)\right)$, the output of $l$-th layer $f^{(l)}$ is referred to an intermediate feature denoted by $\mathcal{F}^{(l)}$.\footnote{We define $\mathcal{F}^{(0)}:=\mathcal{P}$ for the generality of notation,} Our attribution analysis only focuses on those intermediate features representing each point by an individual vector, which are named as \textit{point features}, namely, $\mathcal{F}^{(l)}=\{\mathbf{x}^{(l)}_i\}_{i=1}^N$for $ \mathbf{x}^{(l)}_i \in \mathbb{R}^d$ with a feature dimension $d$. Let $S_f(\mathcal{P})$ denote a scoring function designed to evaluate the model output $f(\mathcal{P})$, the attribution analysis toward the points features $ \mathcal{F}^{(l)}$ is given by 
\begin{equation}
\footnotesize
    r_i^{(l)} = \left\| \frac{\partial S_f(\mathcal{P})}{\partial \mathbf{x}^{(l)}_i}\right\|_2,
\end{equation}
where the term $r_i^{(l)}$ refers to the change of $S_f(\mathcal{P})$ that can be obtained by perturbing the $l$-th layer feature of the $i$-th point. Similar concepts have been defined differently in various literature, such as class-specific saliency~\cite{simonyan2014deep}, pixel sensitivity~\cite{smilkov2017smoothgrad}, or point saliency~\cite{zheng2019pointcloud}. 

In our study, we define $r_i$ as the \textit{point risk} associated with the $i$-th point, as it represents the potential risk of changing the point feature, which could subsequently alter the model output. The set of these terms for all points is referred to as $\mathcal{R}^{(l)}$.  The final point risks are obtained by averaging the normalized risks of multiple point features, given by
\begin{equation}\label{eq:final_risk}
\footnotesize
        r_i = \frac{1}{L'}\sum_{l=0}^{L'} \frac{r_i^{(l)}}{\left\| \nabla_{\mathcal{F}^{(l)}} S_f(\mathcal{P})\right\|_F}.
\end{equation}
where $L'$ is the number of points feature layers and $r_i^{(l)}$ is the unnormalized risk which is then normalized by the Frobenius norm of gradients of the $l$-th layer feature. The motivation to utilize the multi-layer point feature stems from the efficacy of detecting noise when cascading detectors of several intermediate features~\cite{li2017adversarial}.

\begin{figure}[ht!]
  \centering
\includegraphics[width=0.95\linewidth]{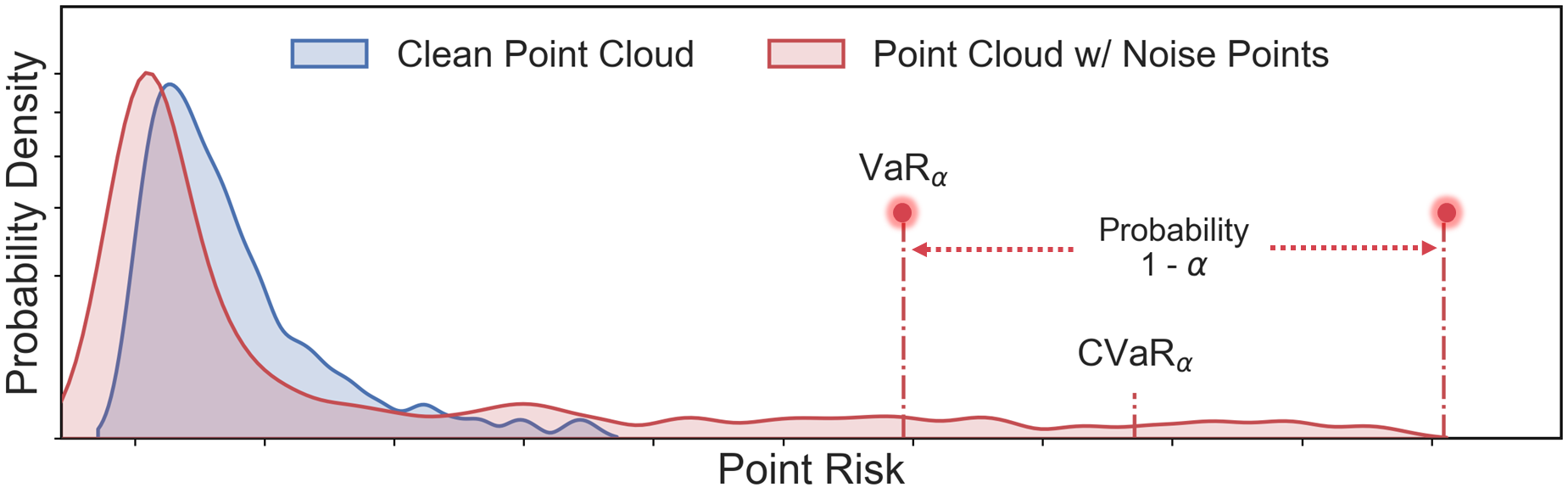}
  \caption{Illustration of CVaR\&VaR on risk distribution.}\label{fig:dist}
\end{figure}

\textbf{Risk Analysis of Outliers.} We calculate and illustrate the point risk distribution of point clouds with or without outliers.  In general, the point cloud with outliers $\mathcal{P}'$ can be formulated from the clean one $\mathcal{P}$ as 
\begin{equation}\label{eq:outlier_format}
\mathcal{P}'\subseteq \mathcal{P}\cup \mathcal{O},
\end{equation}
where $\mathcal{O}$ is a set of noise points serving as the outliers. We note that this is a common setting and is easily realized, thus threatening the real safety, as detailed in Section 4.3.
A typical point risk distribution, as shown in Figure~\ref{fig:intro} and Figure~\ref{fig:dist}, exhibits a long-tail effect, where most points have low risk while only a few have significant risk. This is consistent with previous works \cite{qi2017pointnet, zheng2019pointcloud}, which suggested the uneven contributions of different points to classification. Furthermore, by analyzing the risk distributions, we make the following hypothesis: 
\begin{myhypo}\label{hypo:1}
\small
The presence of noise outliers within a point cloud will raise the level of tail risk in its point risk distribution.
\end{myhypo}
where the tail risk refers to the risks of events accruing rarely but posing a high impact. We believe that this hypothesis is reasonable from two aspects:
First, for the model trained on clean point clouds, the noise points, especially those with attack purposes, are potential outliers, tending to affect the model output. Second, previous works in 2D have verified a similar effect by indicating that noise in samples can be amplified in deep models \cite{lin2019defensive, liao2018defense}. To evidence our hypothesis, we present a numerical measure of tail risks.
\begin{figure*}[ht!]
  \centering
  \includegraphics[width=0.90\linewidth]{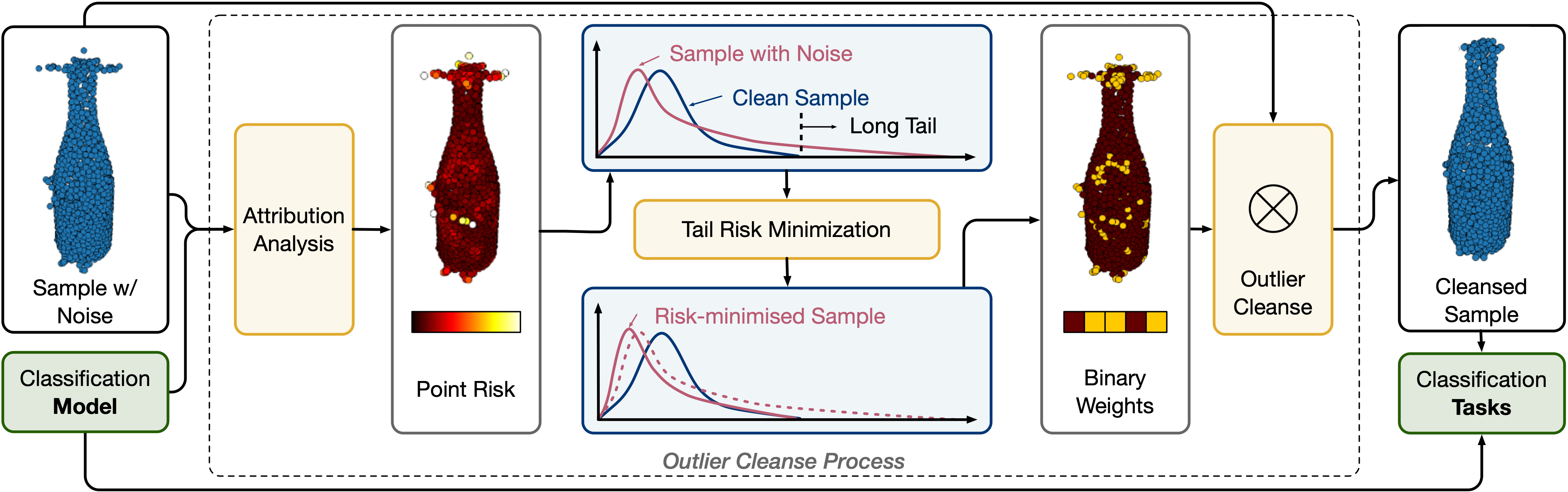}
  \caption{The proposed framework of outlier removal by PointCVaR. Point risks are obtained by entering the noise sample into a trained classification model. Subsequently, an optimization problem is solved to minimize the tail of risk distribution, which leads to binary weights for noise point removal. The processed point cloud is utilized for classification.}\label{fig:method}
\end{figure*}
  
\subsection{Conditional Value-at-Risk}
Conditional Value-at-Risk (CVaR) is a risk measure that quantifies the expected risk that exceeds a certain confidence level, which provides an accurate picture of tail risk~\cite{artzner1999coherent}. To calculate CVaR, we first determine the Value-at-Risk (VaR), which is the maximum risk within a given probability level. Let $R$  be the random variable under a point risk distribution $P$, the definition for VaR is given by
\begin{equation}
\footnotesize
    \text{VaR}_{\alpha}(R) = \inf \left\{\gamma:P(R\leqslant \gamma )\geqslant \alpha \right\},
\end{equation}
where $\alpha$ is the probability level. CVaR is then defined as the expected risk beyond the VaR$_{\alpha}$ level, namely,
\begin{equation}
\footnotesize
    \text{CVaR}_{\alpha}(R) = \mathbb{E}(R \mid R\geqslant \text{VaR}_{\alpha}(R)),
\end{equation}
where $\mathbb{E}(\cdot | \cdot)$ denotes the conditional expectation. An illustration of CVaR is shown in Figure~\ref{fig:dist},  showing that it represents the expectation of long-tailed area within a specific level $\alpha$, \eg, 0.99. A discrete CVaR for point risks is considered in this work. Given the set of risk $\mathcal{R}=\{r_i\}_{i=1}^N$ under a discrete risk distribution $\hat{P}$, the discrete VaR and CVaR are given by
\begin{equation}
\footnotesize
\text{VaR}_{\alpha}(\mathcal{R})=\min\{r_i\in \mathcal{R}: \sum_{r_j\in\mathcal{R}}\mathds{1}(r_i\geqslant r_j)\hat{P}(r_j) \geqslant \alpha\},
\end{equation}
\begin{equation}
\footnotesize
\text{CVaR}_{\alpha}(\mathcal{R}) = \frac{1}{1-\alpha}\sum_{r_i \in \mathcal{R}}\mathds{1}(r_i\geqslant \text{VaR}_{\alpha}(\mathcal{R}))\hat{P}(r_i)r_i.
\end{equation}
where $\mathds{1}(\cdot)$ is the indicator function and $\hat{P}$ is generally set to the empirical distribution, \ie,  $\hat{P}(r_i)=\frac{1}{N}$. With CVaR measuring the tail risk,  we perform comprehensive statistical tests in Section~\ref{sec:section5.3}. Our findings indicate, with a high degree of confidence, that the CVaR value for a clean sample is inferior to that of a noisy sample. Therefore, the second part of Hypothesis~\ref{hypo:1} can be evidenced.

Based on the hypothesis, we suggest that \textbf{it is possible to remove the noise points by minimizing the tail risk of the point clouds}. Furthermore, we propose to directly use CVaR as an objective to optimize the removal process. This is because CVaR possesses the properties of convexity and monotonicity, making it widely used in optimization problems such as risk management~\cite{pflug2000some}. 

\section{Point Cloud Outlier Removal}
In this section, we present an outlier removal method based on point risk optimization, namely, the minimization of Conditional Value at Risk (CVaR). We also provide detailed insights into its practical implementation.

\subsection{Outlier removal by CVaR Optimization}
The outlier removal problem based on Formulation~\eqref{eq:outlier_format} is to remove the potential outliers $\mathcal{O}'$ from $\mathcal{P}'$ thus restore $\mathcal{P}$ as much as possible. Let $\mathcal{R}$ denote the risks of noisy points $\mathcal{P}'$ and $\hat{\mathcal{R}}$ denote the risks of retained points after removal, \ie $\mathcal{P}'\backslash \mathcal{O}'$, the outlier removal with the objective of CVaR minimization is indeed a set selection problem given by
\begin{equation}\label{eq:cvar_opt0}
\footnotesize
\min_{\hat{\mathcal{R}}\subseteq \mathcal{R}} \quad \text{CVaR}_{\alpha}(\hat{\mathcal{R}})\quad  \text{s.t.} \Hquad \big|\hat{\mathcal{R}}\big|\geqslant N_r,
\end{equation}
where $N_r$ is the minimum number of retained points. However, since the point removal will induce the change of model output thus the shift of risk values and distribution $\hat{P}$, this minimization problem is non-trivial. To simplify the problem, we assume the risk values and $\hat{P}$ are \textit{approximately stationary} during the removal, which is theoretically guaranteed by the smoothness assumption of $S_f$ and a large retention ratio, \ie, $\delta=N_r\backslash N\approx 1$. More analysis can be found in Appendix. The problem could also be transformed into a binary optimization
\begin{align}\label{eq:cvar_opt1}
\footnotesize
    \min_{w_1, \cdots, w_N} \quad &\text{CVaR}_{\alpha}(\{w_1r_1, \cdots, w_Nr_N\}), \\ 
    \text{s.t.} \quad &\sum_{i=1}^Nw_i\geqslant N_r,
    \Hquad w_i \in \{0, 1\} \Hquad \forall i\in\{1,\cdots, N\}, \notag 
\end{align}
where $w_i$ is the weight of $i$-th point. A point is retained when its corresponding weight is 1 and removed when it is 0. Subsequently, this optimization can be rendered into a tractable linear programming (LP) problem. This is achieved by two procedures. Firstly, we leverage the conclusions drawn in~\cite{rockafellar2000optimization} and introduce a real additional variable $\zeta$ to obtain the equivalent objective of Equation~\eqref{eq:cvar_opt1}, that is
\begin{equation}
\footnotesize \min_{ \zeta,w_1, \cdots, w_N} \zeta+\frac{1}{1-\alpha}\sum_{i=1}^N\hat{P}(r_i)\left[w_ir_i-\zeta\right]^+.
\end{equation}
Next, we apply linear relaxation to the binary constraints, ultimately converting the problem into the following format
\begin{align}\label{eq:cvar_opt}\footnotesize
    \min_{\substack{\zeta, w_1, \cdots, w_N\\z_1, \cdots, z_N}}\quad&\zeta + \frac{1}{1-\alpha}\sum_{i=1}^N  z_i\hat{P}(r_i), \\ 
    \text{s.t.\ } \quad &\sum_{i=1}^N w_i \geqslant N_r, \quad 1\geqslant w_i\geqslant 0,\\\notag
    &z_i\geqslant w_ir_i-\zeta, \quad z_i\geqslant 0,\quad \forall i \in \{1,\dots,N\}
    .\label{eq:cvar_opt2}
\end{align}
where $z_1, \cdots, z_N$ are the auxiliary decision variables. During minimization, we set a retention rate to approximately 1 for the stationariness approximation of risks, \eg, $\delta=0.95$.  Finally, we set the largest $N_r$ weights in the solution at 1 and the rest at 0 for the removed points; thus, the set selection problem in Equation~\eqref{eq:cvar_opt0} for outlier removal can be tractable.  We refer to this process as \textit{PointCVaR}.

\subsection{Implementation Details}

\textbf{Solution Estimation.} 
Although the LP problem of outlier removals in Equation~\eqref{eq:cvar_opt} is tractable, it often requires polynomial computational complexity in practice~\cite{karmarkar1984new}. To reduce the cost of outlier removal, we employ a risk sorting method to estimate the solution to the problem when $\hat{P}$ is the empirical distribution, which only requires a complexity of $O(N)$. The complete algorithm can be found in Algorithm~\ref{algo:clean}. We note that the estimation is accurate enough in practice, of which a throughout analysis and experimental validation are presented in Appendix.
\begin{algorithm}
 \caption{Outlier Removal by PointCVaR}
 \begin{algorithmic}[1]\label{algo:clean}
 \renewcommand{\algorithmicrequire}{\textbf{Input:}}
 \renewcommand{\algorithmicensure}{\textbf{Output:}}
 \REQUIRE A model $f$, retention rate $\delta$,  scoring function $S_f$\\
a noisy point cloud $\mathcal{P}'=\{\mathbf{x}'_i\}_{i=1}^N$ .
 \ENSURE Processed point cloud $\mathcal{P}'\backslash\mathcal{O}'$
  \STATE $\mathcal{R}\leftarrow $ Get $S_f( \mathcal{P}')$ and perform Equation~\eqref{eq:final_risk} on $\mathcal{P}'$.
  \STATE $r_t \leftarrow \min\{r_i\in\mathcal{R}: \sum_{r_j \in\mathcal{R}} \mathds{1}(r_i\geqslant r_j)\geqslant \lceil \delta N \rceil\}$.
\STATE $\mathcal{O}'\leftarrow \emptyset $.
\FOR {$i\in \{1,\dots,N\}$}
    \IF{$r_i\geqslant r_t$}
    \STATE $\mathcal{O}'\leftarrow \mathcal{O}' \cup \{\mathbf{x}'_i\}$.
    \ENDIF
  \ENDFOR

  \RETURN $\mathcal{P}'\backslash \mathcal{O}'$.
 \end{algorithmic}
 \end{algorithm}

\begin{table*}[ht!]
\footnotesize
\centering
\begin{tabular}{l|rr|rrrr|rr|rr|rrrr|rr}
\toprule[1pt]
\multirow{2}{*}{Methods} & \multicolumn{2}{c|}{Rand (MN)}                      & \multicolumn{4}{c|}{Adv (MN)}                                                                         & \multicolumn{2}{c|}{BA (MN)}                      & \multicolumn{2}{c|}{Rand(SN)}                      & \multicolumn{4}{c|}{Adv (SN)}                                                                         & \multicolumn{2}{c}{BA (SN)}                      \\
                         & \multicolumn{1}{c}{Glb} & \multicolumn{1}{c|}{Loc} & \multicolumn{1}{c}{HD} & \multicolumn{1}{c}{CD} & \multicolumn{1}{c}{Obj} & \multicolumn{1}{c|}{Cls} & \multicolumn{1}{c}{PL} & \multicolumn{1}{c|}{CL} & \multicolumn{1}{c}{Glb} & \multicolumn{1}{c|}{Loc} & \multicolumn{1}{c}{HD} & \multicolumn{1}{c}{CD} & \multicolumn{1}{c}{Obj} & \multicolumn{1}{c|}{Cls} & \multicolumn{1}{c}{PL} & \multicolumn{1}{c}{CL} \\ \hline\hline
Raw                      & 5.4                     & 52.6                     & 0.9                    & 2.4                    & 0.4                     & 0.3                      & 4.1                    & 39.2                    & 11.0                    & 73.8                     & 1.7                    & 3.7                    & 2.6                     & 4.1                      & 10.0                   & 34.7                   \\
RS                       & 8.4                     & 52.6                     & 73.7                   & 77.9                   & 30.1                    & 45.1                     & 4.1                    & 36.8                    & 12.2                    & 74.6                     & 84.2                   & 83.0                   & 43.4                    & 57.4                     & 10.0                   & 33.0                   \\
ROR                      & 82.1                    & 64.7                     & 74.5                   & 81.9                   & 35.0                    & 59.4                     & 4.1                    & 35.3                    & 97.2        & 83.6                     & 88.1                   & 93.8                   & 47.2                    & 72.1                     & 10.0                   & 34.5                   \\
SOR                      & 72.3                    & \underline{65.7}         & 82.7                   & 84.9                   & 36.9                    & 63.9                     & 4.1                    & 37.3                    & 78.0                    &  \underline{84.3}                     & 95.8                   & 96.0                   & 52.2                    & 75.2                     & 10.0                   & 34.6                   \\
PCN                  & 45.6                    & 56.4                     & 69.8                   & 57.0                  & 48.2                    & 61.9                     & 4.1                    & 38.1        & 46.3                    &78.6         & 43.8                   & 38.7                   & 34.1                    & 35.9                     & \underline{15.6}                   & 36.4                   \\
\rowcolor{Gray} Vanilla                  & \underline{82.4}        & 63.9                     & \underline{86.2}          & \underline{88.7}          & \underline{48.9}        & \underline{63.9}         & \underline{4.7}        & \underline{39.6}        & \underline{97.6}        & 82.7                     & \underline{97.4}          & \underline{97.3}          & \underline{61.4}        & \underline{76.7}         & 10.0        & \underline{37.2}       \\
\rowcolor{Gray}Multistep                    & \underline{88.0}             & \underline{70.0}              & \underline{84.0}         & \underline{85.3}       & \underline{62.8}           & \underline{67.8}            & \underline{85.7}          & \underline{78.3}           & \underline{97.7}           & \underline{85.5}              & \underline{97.3}       & \underline{96.2}       & \underline{67.5}           & \underline{79.0}              & \underline{88.9}          & \underline{75.4}          \\ \bottomrule[1pt]
\end{tabular}
\caption{Outlier removal method comparison based on PointNet accuracy (\%) on ModelNet40 (MN) and ShapeNet (SN) corrupted by different outliers, where the top two accuracies are highlighted by underlines.} \label{tab:main}
\end{table*}

\textbf{Scoring Function.} 
We introduce the scoring function to calculate the risk of points, which is divided into two terms. The first term is the classification score, which employs the commonly used cross-entropy loss on the model prediction. This is denoted by $S_c(f(P))$. The second term is the geometric score, for which we incorporate a differentiable score that measures the uniformity of point clouds. The formula is
\begin{equation}
\footnotesize
S_g(\mathcal{P}) = \sqrt{\frac{1}{N}\sum_i^N (d_i-\bar{d})^2},\\
\end{equation}
\begin{equation}
\footnotesize
d_i=\frac{1}{|\mathcal{N}_i|}\sum_{\mathbf{x}_j\in \mathcal{N}_i} \left\|\mathbf{x}_i-\mathbf{x}_j\right\|_2,\Hquad \bar{d} = \frac{1}{N}\sum_i^N d_i,
\end{equation}
where $\mathcal{N}_i\subseteq\mathcal{P}$ refers to the nearest neighbors of $x_i\in\mathcal{P}$. Therefore, $d_i$ measures the neighborhood distance of a point, and $S_g(\mathcal{P})$  is the variance of $d_i$. The final scoring function is given by $S_f(\mathcal{P})=S_c(\mathcal{P}, f)+\lambda S_g(\mathcal{P})$ with a tunable parameter $\lambda$. It is noted that the geometric score only affects the risks of input points rather than other point features.

\textbf{Multi-step Outlier Removal.} 
In order to better satisfy the condition of approximate stationariness of risks for the LP formulation, we propose a multi-step removal method where the retention rate of each step is closer to 1 than $\delta$, Specifically, it is achieved by executing Algorithm~\ref{algo:clean} by $I$ times. The retention rate of each step is $\delta^{\frac{1}{I}}$. Although this results in a linear increase in computational complexity, experiments suggest that it surpasses single-step removal due to a superior approximation.  In subsequent discussions, we refer to the multi-step method and the original single-step method as \textit{multi-step} and \textit{vanilla}, respectively.

\section{Experiment}
\subsection{Experiment Settings}
\textbf{Dataset and Model.} Our experiments are conducted based on ModelNet40~\cite{wu20153d} (MN) and ShapeNet~\cite{shapenet2015} (SN) for classification task. The train$\backslash$test splits of them follow the implementation of PointNet~\cite{qi2017pointnet}, which are 9843$\backslash$2468 for MN and  12128$\backslash$2874 for SN. Each clean point cloud in the dataset contains 1024 points while this number may vary for the noisy data. For the models, we utilize PointNet, DGCNN, PointCloudTransformer (PCT) and GDANet by following the official implementations~\cite{qi2017pointnet, wang2019dynamic, guo2021pct, xu2021learning}. Unless otherwise specified, we perform experiments on the models trained by the protocols as detailed by~\cite{goyal2021revisiting}.

\textbf{Noise Setting.} 
We implement three types of point cloud outliers that can occur in real-world scenarios, which are:
\begin{itemize}[leftmargin=*]
    \item\textbf{Random Outlier (Rand)} comprises of random global (Glb) and local (Loc) noise that imitate background points in real-world data~\cite{ren2022benchmarking}, respectively. 
    \item \textbf{Adversarial Outlier (Adv)} is created by implementing the point addition attacks listed in~\cite{xiang2019generating}, which includes adding points by Hausdorff distance (HD), Chamfer divergence (CD), objects (Obj), and clusters (Cls) methods. Studies have demonstrated that such attacks can be physically realizable~\cite{tu2020physically}.
    \item \textbf{Backdoor Trigger (BA)} involves a small sphere located at a specific coordinate, as proposed in~\cite{li2021pointba}. This outlier can impact the models trained dataset corrupted by poison-label (PL) or clean-label (CL) methods, which can be easily generated in the real world~\cite{li2021pointba}.
\end{itemize}
We add the above noises into the data with more details including hardware configurations in Appendix. Moreover, we also evaluate the effectiveness of combining existing methods and ours on resisting various adversarial noises including perturbation attacks by C\&W (CW)~\cite{xiang2019generating}, $k$NN~\cite{tsai2020robust}, shape-invirant (SI)~\cite{huang2022shape} and AdvP (APC)~\cite{hamdi2020advpc} methods, as well as the point drop attack~\cite{zheng2019pointcloud}.

\textbf{Parameters Setting.} 
We set $\delta$ to be 0.95 and 0.92 for the multi-step and vanilla removal method, $\lambda=1.0$ for the scoring function, and $I=20$ for the multi-step method. Further investigations into the effects of these parameters are presented in Section~\ref{sec:section5.3}.

\textbf{Evaluation Protocol.} 
During the training phase, we train the model on either a clean dataset or one that is backdoor-corrupted. In the testing phase, we first perform an outlier removal on the noisy test set. Subsequently, we report the test accuracy of the model on the processed data. A higher accuracy is indicative of superior performance of the removal method. We refer the protocol as removal-and-classification.

\subsection{Main Results}
\textbf{Removal and Classification.}
We conduct extensive tests for our outlier removal method based on different noisy versions of ModelNet40 and ShapeNet. For better comparison, we also implement three outlier removal methods: Random Sampling (RS)~\cite{yang2019adversarial}, Radius Outlier Removal (ROR)~\cite{rusu20113d}, and SOR~\cite{rusu2010semantic}. Additionally, we re-implement the outlier removal network in PointCleanNet (PCN)~\cite{rakotosaona2020pointcleannet} as a baseline for learning-based methods.
We perform these methods based on the proposed parameters and training process, while for ROR, we set $r_{ror}=0.1$ and $n_{ror}=4$ as detailed in Appendix. The results in Table~\ref{tab:main} and Table~\ref{tab:main2} show that our multi-step method consistently outperforms all baseline methods, particularly as the only method that effectively removes backdoor triggers achieving above 80\% accuracy for backdoor attacks across all models and datasets. Furthermore, the vanilla method serves as a cost-effective removal option, exhibiting decent performance on all models at a relatively low computational cost. In particular, its execution speed rivals that of SOR, surpassing PCN by more than 10 times, even without considering the training time of PCN. More runtime profiles are provided in Appendix.

\begin{table}[h!]
\centering
\resizebox{0.46 \textwidth}{!}{
\begin{tabular}{ccrrrrr>{\columncolor{Gray}}r>{\columncolor{Gray}}r}
\toprule[1pt]
Noise                                                               & Model  & \multicolumn{1}{c}{Raw} & \multicolumn{1}{c}{RS} & \multicolumn{1}{l}{ROR} & \multicolumn{1}{c}{SOR} & \multicolumn{1}{c}{PCN} & \multicolumn{1}{c}{V} & \multicolumn{1}{c}{M} \\ \hline\hline
\multirow{3}{*}{\begin{tabular}[c]{@{}c@{}}Glb\\ (MN)\end{tabular}} & DGCNN  &65.7                     & 52.2                   & 83.6                     & 87.4                     & 75.2                    & \underline{89.1}                   & \underline{89.6}                   \\
                                                                    & PCT    & 49.7                     & 52.6                   & 83.3                    & 85.9                    & 69.2                   & \underline{88.7}                  & \underline{90.2}                  \\
                                                                    & GDANet & 63.7                   & 51.6                   & 82.7                    & 88.3                    &74.0                    & \underline{88.6}                  & \underline{90.3}                  \\ \hline
\multirow{3}{*}{\begin{tabular}[c]{@{}c@{}}CD\\ (MN)\end{tabular}}  & DGCNN  & 1.5                    & 76.9                   & 69.1                    & 73.0                    &71.8                    & \underline{84.0}                  & \underline{80.0}                  \\
                                                                    & PCT    & 0.4                    & 84.4                   & 82.6                    & 85.1                    &79.4                    & \underline{88.5}                  & \underline{87.3}                  \\
                                                                    & GDANet & 3.8                    & 78.8                   & 63.6                    & 74.6                    &71.7                    & \underline{83.9}                 & \underline{82.2}                  \\ \hline
\multirow{3}{*}{\begin{tabular}[c]{@{}c@{}}PL\\ (MN)\end{tabular}}  & DGCNN  & 4.1                    & 61.4                   & 4.1                    & 14.6                    &4.1                    & \underline{83.7}                  & \underline{88.6}                  \\
                                                                    & PCT    & 4.1                     & 80.7                   & 48.1                     & 55.0                     &52.8                     & \underline{83.6}                   & \underline{87.7}                   \\
                                                                    & GDANet & 4.1                     & 78.5                   & 8.8                    & 37.3                    &31.1                    & \underline{87.6}                  & \underline{88.7}                  \\ \hline\hline
\multirow{3}{*}{\begin{tabular}[c]{@{}c@{}}Glb\\ (SN)\end{tabular}} & DGCNN  & 59.1                     & 62.5                   & \underline{97.8}                     & 88.0                     &79.9                     & 96.7                   & \underline{98.1}                   \\
                                                                    & PCT    & 36.5                     & 41.4                   & \underline{98.3}                    & 93.4                    & 71.8                   & 97.6                  & \underline{98.5}                  \\
                                                                    & GDANet & 22.3                    &25.4                    & 97.1                    & 84.7                    &65.6                    & \underline{97.2}                  & \underline{97.8}                  \\ \hline
\multirow{3}{*}{\begin{tabular}[c]{@{}c@{}}CD\\ (SN)\end{tabular}}  & DGCNN  & 4.1                    & 84.4                   & 88.0                    & 92.1                    &75.8                    & \underline{95.6}                  & \underline{93.7}                  \\
                                                                    & PCT    & 0.1                    & 90.1                   & 90.6                    & 94.7                    &72.3                    & \underline{96.4}                  & \underline{95.9}                  \\
                                                                    & GDANet & 1.7                    & 82.4                   & 88.9                    & 92.8                    &68.2                    & \underline{94.8}                  & \underline{94.1}                  \\ \hline
\multirow{3}{*}{\begin{tabular}[c]{@{}c@{}}PL\\ (SN)\end{tabular}}  & DGCNN  & 10.0                    & 88.3                   & 10.0                    & 34.8                    &34.4                    & \underline{99.0}                  & \underline{97.9}                  \\
                                                                    & PCT    & 10.0                     & 92.9                   & 58.1                     & 87.0                     &88.7                     &  \underline{93.5}                   &  \underline{96.1}                   \\
                                                                    & GDANet & 10.0                     & 95.8                   & 16.5                    & 73.1                    &93.6                    & \underline{97.9}                  & \underline{98.8}                  \\ \bottomrule[1pt]
\end{tabular}
}
\caption{Method comparison (V for vanilla and M for multistep) based on DGCNN, PCT and GDANet. Accuracy (\%) on corrupted ModelNet40 and ShapeNet are presented.} \label{tab:main2}
\end{table}

\textbf{Plugin Performance.} As outlier removal is frequently used for point cloud pre-processing~\cite{rusu2010semantic}, we implement our approach as a pre-processing plugin to existing robust point cloud classification methods, thereby extensively improving their robustness to adversarial attacks. The plugin is applied to three methods, DUP-Net, PointCutMix (PCM)~\cite{zhang2022pointcutmix} and PGD adversarial training (PGD)~\cite{sun2021adversarially}. Particularly, we replace SOR with our method in DUP-Net, and also preprocess the input for PCM$\backslash$PGD trained model prior to inference. For lightweight consideration of plugins, we only utilize the vanilla method in the experiments. 
Results in Table~\ref{tab: comp} indicate that the proposed technique consistently improves the performance of DUP-Net, PCM, and PGD by a significant margin, particularly for the latter two model-based methods where they are vulnerable to adaptive attacks. Notably, PGD achieves an average accuracy of 80\% and 90\%  on ModelNet40 and ShapeNet with adversarial noise, respectively. Given the low cost of this method, it demonstrates substantial potential as a plugin module to enhance point cloud classification.
\addtocounter{figure}{1}
\begin{figure*}[ht!]
    \centering
        \includegraphics[width=\linewidth]{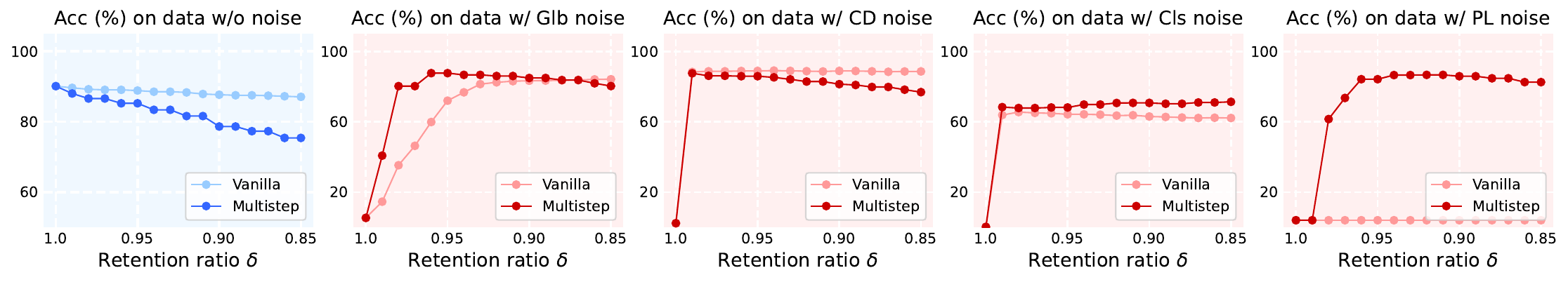}
\caption{Accuracy on ModelNet40 without or with different noise \vs retention rate of PointCVaR.}
\label{fig:four_acc}
\end{figure*}

\begin{table}[h!]
\centering
\resizebox{0.47\textwidth}{!}{\begin{tabular}{ccrrrrrrr}
\toprule[1pt]
Method                & Plugin    & CD                   & Cls
& $k$NN                          & CW                        & \multicolumn{1}{c}{SI}                  & APC                  & Drop                                         \\ \hline\hline
                      & $\times$    & 85.3                 & 63.7                 & 77.1                         & 85.9                        & 40.1                 & 71.9                 & 61.1                                         \\
\multirow{-2}{*}{DN (MN)}  & \checkmark & \cellcolor{Gray}87.3 & \cellcolor{Gray}64.5 & \cellcolor{Gray}81.8         & \cellcolor{Gray}86.2        & \cellcolor{Gray}81.7 & \cellcolor{Gray}74.5 & \cellcolor{Gray}62.7                         \\ \hline
                      & $\times$    & 2.8                  & 4.3                  & 39.2                         & 1.8                         & 2.3                  & 5.3                  &76.2                                              \\
\multirow{-2}{*}{PCM (MN)} & \checkmark & \cellcolor{Gray}89.8 & \cellcolor{Gray}87.2 & \cellcolor{Gray}75.6         & \cellcolor{Gray}89.1        & \cellcolor{Gray}80.3 & \cellcolor{Gray}59.2 & \cellcolor{Gray}76.3                         \\ \hline
                      & $\times$    & 3.8                  & 6.6                  & 45.3                         & 1.9                         & 4.2                  & 0.8                  & 80.3                 \\
\multirow{-2}{*}{PGD (MN)} & \checkmark & \cellcolor{Gray}88.9 & \cellcolor{Gray}80.3 & \cellcolor{Gray}82.6         & \cellcolor{Gray}87.7        & \cellcolor{Gray}78.3 & \cellcolor{Gray}70.6 & \cellcolor{Gray}80.3 \\ \hline\hline
                      & $\times$    & 97.1                 & 74.9                 & 95.6                         & 96.5                        & 45.9                 & 88.7                 & 90.3                                         \\
\multirow{-2}{*}{DN (SN)}  & \checkmark & \cellcolor{Gray}97.1 & \cellcolor{Gray}75.3 & \cellcolor{Gray}95.9         & \cellcolor{Gray}96.8        & \cellcolor{Gray}92.5 & \cellcolor{Gray}93.1 & \cellcolor{Gray}90.6                         \\ \hline
                      & $\times$    & 2.8                  & 4.5                  & 39.2                         & 1.8                         & 2.5                  & 5.3                  & 78.4                                         \\
\multirow{-2}{*}{PCM (SN)} &\checkmark             & \cellcolor{Gray}90.5 & \cellcolor{Gray}85.2 & \cellcolor{Gray}85.7         & \cellcolor{Gray}87.5        & \cellcolor{Gray}88.7 & \cellcolor{Gray}64.8 & \cellcolor{Gray}81.9                         \\ \hline
                     & $\times$    & 4.4                  & 9.7                 & 83.2 & 1.6 & 31.6                 & 33.5                 & 92.6                                         \\
\multirow{-2}{*}{PGD (SN)} & \checkmark & \cellcolor{Gray}97.2 & \cellcolor{Gray}88.9 & \cellcolor{Gray}94.7         & \cellcolor{Gray}90.6        & \cellcolor{Gray}94.5 & \cellcolor{Gray}81.5 & \cellcolor{Gray}93.5                         \\ \bottomrule[1pt]
\end{tabular}
}
\caption{PointNet accuracy (\%) of three robust classification methods on ModelNet40 and ShapeNet with diverse adversarial noises. The comparison is made between the original method and the one enhanced by our Plugin method. } \label{tab: comp}
\end{table}
\addtocounter{figure}{-2}
\begin{figure}[hb!]
    \centering
    \includegraphics[width=0.97\linewidth]{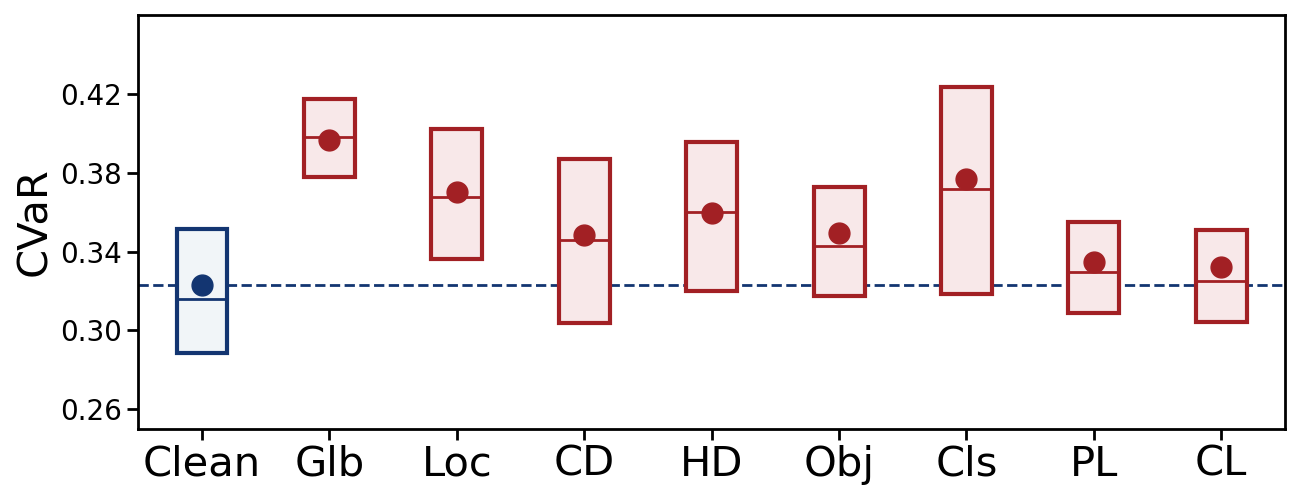}
    \caption{ CVaR$_{0.99}$ boxplots (Mean, median and $25\backslash75$ quantiles) of clean sets risks (blue) and noisy sets risks (red).}
    \label{fig:test}
\end{figure}

\subsection{More Studies} \label{sec:section5.3}
\textbf{Hypothesis Test.}
We conduct the statistical testing for Hypothesis~\ref{hypo:1}, where CVaR$_{0.99}$ is used to measure the tail risk. Figure~\ref{fig:test} showcases that the risks of clean samples generally have lower CVaR$_{0.99}$ compared to the samples with noise points. Furthermore, we use a paired Wilcoxon signed test to investigate the difference between the risk of a clean sample and its noise counterparts. The results showed that all p-values obtained from the one-sided tests are less than 0.01. Therefore, we reject the null hypothesis and conclude that the \textbf{tail risk of a sample with noise is statistically greater than its corresponding clean sample}.

\begin{figure}[hb!]
    \centering
        \hspace{0em}\includegraphics[width=0.98\linewidth]{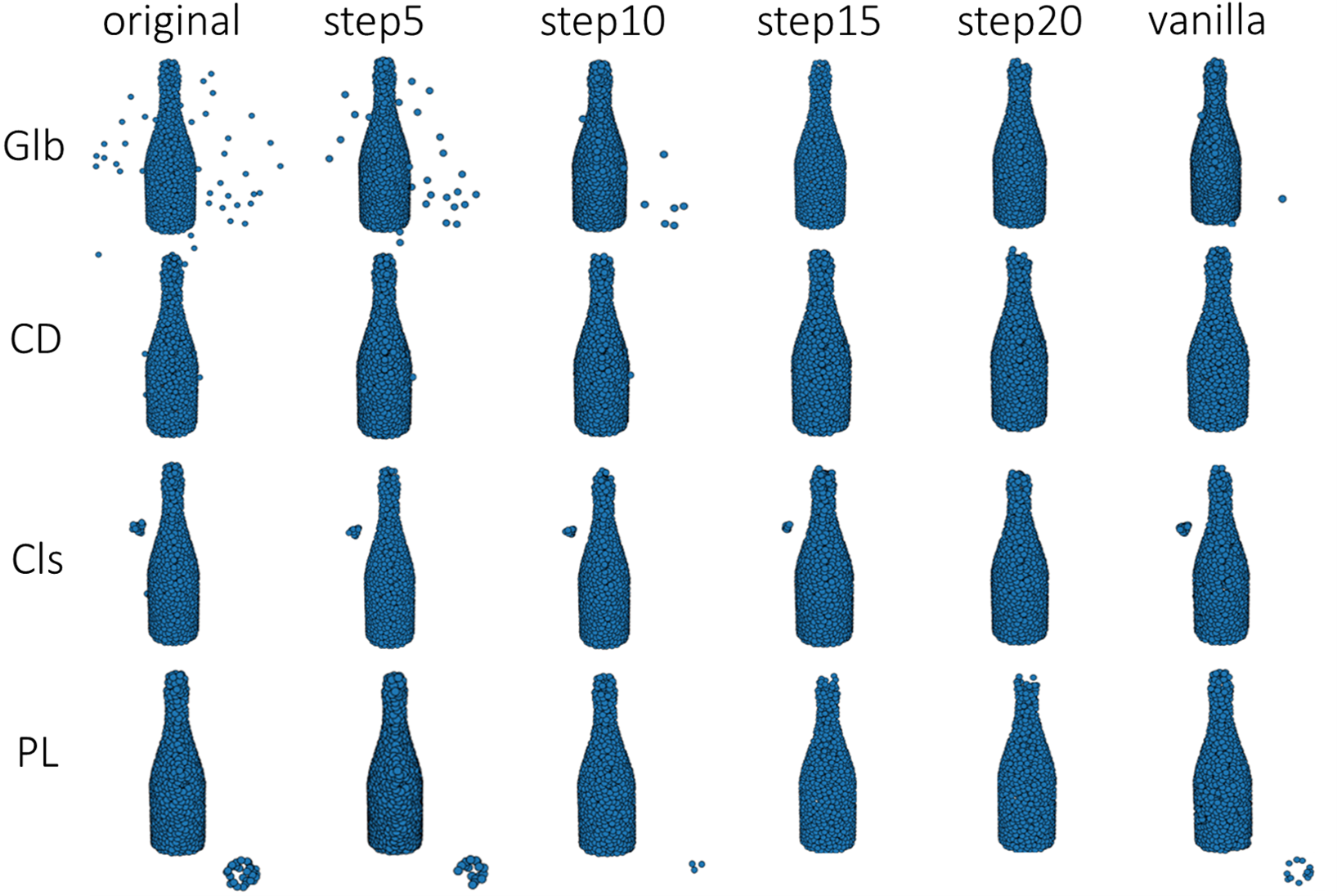}
        \caption{Visualization of PointCVaR removal process.}
        \label{fig:vis_drop}
\end{figure}

\textbf{Effect of Retention Rate.} In Figure~\ref{fig:vis_drop}, the removal process is visually depicted. It is evident that the vanilla method sometimes struggles to eliminate outliers completely, particularly backdoor triggers. Conversely, the multi-step method consistently removes noise. Nevertheless, multi-step tends to over-remove clean points when dealing with outliers of small proportion such as CD. Consequently, we present a further analysis of the impact of the retention rate $\delta$ in Figure~\ref{fig:four_acc}. This figure demonstrates how model accuracy on processed samples changes as $\delta$ decreases. Although it generally outperforms the vanilla method in terms of performance, the multi-step method leads to a more pronounced decrease in accuracy in clean samples when $\delta$ is low.

We hypothesize that this is due to the removal of so-called salient points from clean samples, as suggested in prior work~\cite{zheng2019pointcloud}. However, the decrease in accuracy for clean samples is slower than the increase in accuracy on noisy samples, enabling us to always locate a suitable $\delta$ that yields satisfactory performance. For instance, $\delta=0.95$ for the multi-step method and $\delta=0.92$ for the vanilla method yield accuracies of 85\% and 88.5\% on clean samples, respectively, alongside acceptable accuracy. The differing rates of change can be attributed to the lower tail risks of clean samples compared to noisy ones, which results in greater stability during the removal of high-risk points.

\begin{table}[]
\centering
\resizebox{0.45 \textwidth}{!}{
\begin{tabular}{cccccccc}
\toprule[1pt]
Method   & CSF & $\lambda$ & $I$ & MPF        & \multicolumn{1}{c}{Glb} & \multicolumn{1}{c}{CD } & \multicolumn{1}{c}{PL } \\ \hline\hline
Vanilla  & CE  & 1.0       & N.A. & \checkmark & \underline{82.4} & \underline{88.7}                        & \underline{4.7}                         \\
Vanilla  & Max & 1.0       & N.A. & \checkmark & 76.7                         & 88.5                        & 4.1                         \\
Vanilla  & Sum & 1.0       & N.A. & \checkmark & 77.2                         & 88.3                        & 4.1                         \\
Vanilla  & CE  & 0.0       & N.A. & \checkmark & 59.5                         & 88.5                        & 4.1                         \\
Vanilla  & CE  & 1.0       & N.A. & $\times$   & 81.4                         & 88.3                        & 4.3                         \\\hline
Multstep & CE  & 1.0       & 5    & \checkmark & \underline{88.0}                         & \underline{86.9}                        & 35.8                        \\
Multstep & CE  & 0.0       & 20   & \checkmark & 87.2                         & 85.5                        & 84.2                        \\
Multstep & CE  & 1.0       & 20   & $\times$   & 86.8                         & 86.3                        & 78.1                        \\
Multstep & CE  & 1.0       & 20   & \checkmark & 88.0                         & 85.3                        & \underline{85.7}                        \\ \bottomrule[1pt]
\end{tabular}}\caption{Ablation study based on the accuracy of data with global, add CD and backdoor triggers. CSF and MPF represent the classification scoring function $S_c$ and the use of multi-layer point features for risk calculation, respectively. CE is the cross-entropy loss, while the Max and Sum are the maximum and sum of output logits, respectively.}\label{tab:ablation}
\end{table}

\textbf{Ablation Study.} We conduct ablation studies on various elements within our outlier removal approach. As demonstrated in Table 4, the selection of the classification score $S_c$ has a minimal impact on our method, only causing a slight performance variation in the vanilla method's handling of global noise. On the other hand, the introduction of the geometric score $S_g$ 
 , \ie, $\lambda=1$, consistently enhances our method's capacity to remove global noise, without affecting other types of noise. Also, our findings indicate that, compared to the vanilla method, the multi-step method exhibits less sensitivity to changes in these components. The only factor substantially affecting the performance of the multi-step method is the number of iterations $I$, as a successful removal of backdoor triggers necessitates a higher $I$.

\section{Conlusion}
In conclusion, the noise effect has cast a shadow over the reliability of 3D deep learning on point clouds in industrial applications. Our proposed outlier removal method leverages trained classification models to effectively filter out many types of outliers in point clouds. By utilizing gradient-based attribution analysis and tail risk minimization, we formulate the process into an optimization problem and use linear programming to obtain binary weights for input points. This method consistently improves the classification accuracy of deep models on different versions of noisy point clouds. It can also be combined with other methods to defend against a larger range of adversarial noises at a low cost.

\bibliography{arxiv}

\end{document}